\documentclass[conference]{IEEEtran}
\ifCLASSINFOpdf
  \usepackage[pdftex]{graphicx}
  \graphicspath{{./pic/}}
  \DeclareGraphicsExtensions{.pdf,.jpeg,.png}
\else
  \usepackage[dvips]{graphicx}
  \graphicspath{{../eps/}}
  \DeclareGraphicsExtensions{.eps}
\fi
\usepackage[table,xcdraw]{xcolor}
\usepackage{adjustbox}
\usepackage{multirow}
\usepackage{amsmath}
\usepackage{amssymb}
\newcommand{\abs}[1]{ \left\lvert#1\right\rvert}

\usepackage{algorithm}
\usepackage{algpseudocode}
\usepackage{etoolbox}
\makeatletter
\patchcmd{\@makecaption}
  {\scshape}
  {}
  {}
  {}
\makeatother

\floatname{algorithm}{Procedure}

\begin{document}
%
\title{Computation-Performance Optimization of Convolutional Neural Networks with Redundant Kernel Removal}



%
\author{\IEEEauthorblockN{Chih-Ting Liu,
Yi-Heng Wu,
Yu-Sheng Lin, and 
Shao-Yi Chien}
\IEEEauthorblockA{Media IC and System Lab\\
Graduate Institute of Electronic Engineering and Department of Electrical Engineering\\
National Taiwan University\\
No. 1, Sec. 4, Roosevelt Road, Taipei, 10617 Taiwan}
\IEEEauthorblockA{{\tt\small \{jackieliu, jasonwu, johnjohnlys, sychien\}@media.ee.ntu.edu.tw}}} 


\maketitle

\begin{abstract}  
Deep Convolutional Neural Networks (CNNs) are widely employed in modern computer vision algorithms,
where the input image is convolved iteratively by many kernels to extract the knowledge behind it.
However, with the depth of convolutional layers getting deeper and deeper in recent years,
the enormous computational complexity makes it difficult to be deployed on embedded systems with limited 
hardware resources.
In this paper, we propose two computation-performance optimization methods to reduce the redundant convolution kernels of a CNN with performance and architecture constraints,
and apply it to a network for super resolution (SR).
Using PSNR drop compared to the original network as the performance criterion,
our method can get the optimal PSNR under a certain computation budget constraint.
On the other hand, our method is also capable of minimizing the computation required under a given PSNR drop. 
\end{abstract}


%
\IEEEpeerreviewmaketitle

\section{Introduction}   
In recent years, deep neural network has been widely employed in the state-of-the-art works in many fields like computer vision \cite{krizhevsky2012imagenet}, natural language processing \cite{collobert2011natural}, and speech recognition \cite{hinton2012deep}. Convolutional Neural Networks (CNNs) are getting a great success especially in many visual tasks, including image classification \cite{simonyan2014very}, object detection \cite{girshick2014rich}, image style transfer \cite{gatys2016image}, super resolution \cite{kim2016accurate}, etc. 
Though these CNNs are powerful, they often consume substantial storage and computational resources. Commonly, the training step of CNNs can be carried out offline on high performance CPU/GPU clusters;
nevertheless, for the inference stage, we often prefer local computation on embedded systems rather than cloud-based solutions owing to the privacy, latency, transmission bandwidth, and power consumption constraints \cite{chen2016deep}.
Therefore, reducing the parameters in CNNs in order to avoid huge storage and computation complexity has become critical.

Several works have been proposed to reduce the redundancy of neural networks.
Network pruning based on the weight magnitude can remove insignificant connections between neurons \cite{han2015learning,han2015deep}.
\cite{jaderberg2014speeding} shows that the redundancies in CNN convolutional layers can be exploited by approximating the filter bank as combinations of a rank-1 filter bias. 
Several works optimize neural networks by quantizing the weights and intermediate results of CNNs 
\cite{vanhoucke2011improving,hung2015,anwar2015}. 
Vector quantization is also been used in compressing the weighting parameters as well as reducing the computational complexity \cite{gong2014compressing, wu2016quantized}.
However, when deploying these redundancy removal methods on existing embedded devices, the performance improvement highly depends on the processor architecture and the effort to optimize the implementation accordingly.
A kernel pruning approach \cite{chen2016deep} removes kernels with high sparsity to reduce the computation.
One advantage of this approach is that it is easy to be deployed on all kinds of computing architectures.

In this paper, two computation-performance optimization (CPO) methods are proposed for constrained computing environment, such as embedded systems. Based on \cite{chen2016deep}, redundant kernels are removed according to the computation resource limitation and performance (quality or accuracy) requirements. 
The experiment model is a network for super resolution (SR) in \cite{kim2016accurate}, and the performance benchmark is the PSNR drop compared to the original SR network.
In the first method, layer-wise polishment (LWP), under a specified computational budget, the minimal PSNR drop is achieved, where we sort the kernels by its sparsity and modify the removal percentage of every layer.
Second, under a given PSNR drop requirement, the minimized computation is achieved by gradient optimization (GO), where a regression model is trained for optimizing the kernel removal percentage.
Specifically, our contributions are:
\begin{enumerate}
\item Applying the \textit{kernel redundancy removal} method on a super resolution (SR) network,
\item Adjusting the reducing factor of each layer by kernel sparsity, optimizing the PSNR drop under a computation budget and
\item Generating an appropriate reducing factor to optimize the network with our trained regression model under given PSNR drop.
\end{enumerate}

\section{Proposed Method}  
The opreations of a CNN layer involve convolving a 3-D tensor (input, $\mathbf{X} \in \mathbb{R}^{C\times Y\times X}$) with 4-D tensors (kernels, $\mathbf{W} \in \mathbb{R}^{N\times C\times H\times W}$) to extract different features and then generating a 3-D tensor (output, $\mathbf{Y} \in \mathbb{R}^{N \times Y^{\prime} \times X^{\prime}}$), where $C,Y,X$ are channel, height and width of input tensor, $C,W,H$ are channel, height and width of one convolution kernel, and $N$ is the number of convolution kernels. $Y^{\prime}, X^{\prime}$ are slightly different from $Y, X$ owing to the boundary of convolution.
The output tensor is also the input tensor of the next layer.
In this paper, we focus on the 4-D convolutional kernels and propose two methods to prune the redundant kernels layer by layer, achieving computation-performance optimization (CPO).

\subsection{Definition of Redundancy}
The criterion of redundancy is defined layer-by-layer according to the weights in a layer. For a specified layer $l$, We use $M_l$ to represent the mean value of all absoluted kernel weights at that layer:
\begin{equation}
    M_l = \frac{\sum_{n,c,h,w}\abs{k_{l, nchw}}}{N\times C\times W \times H},
\end{equation}
where the $n,c,h,w$ are the indices of the 4-D tensor.
Then the sparsity of the $n$-th kernel $S_l(n)$ at layer $l$ can be written as:
\begin{align}
    \begin{split}
        S_l(n) = \frac{\sum_{c,h,w}\sigma (k_{l, nchw})}{C\times W \times H} \\
    \sigma(x) = \left\lbrace\begin{aligned}
        1&, \text{if} \abs{x} < M_l  \\ 
        0&, \text{otherwise}  \label{eq:sparsity}
    \end{aligned}\right.
\end{split}
\end{align}
In other words, if a kernel has several coefficients which are less than the mean value at that layer, $S_l(n)$ is close to $1$, which means this kernel is more redundant than others.
Based on this representation, our two proposed CPO methods, \textit{Layer-Wise Polishment} (LWP) and \textit{Gradient Optimization} (GO), are described as follows. 
 
\begin{figure}[!t]  
\centering
\includegraphics[width=3.3in]{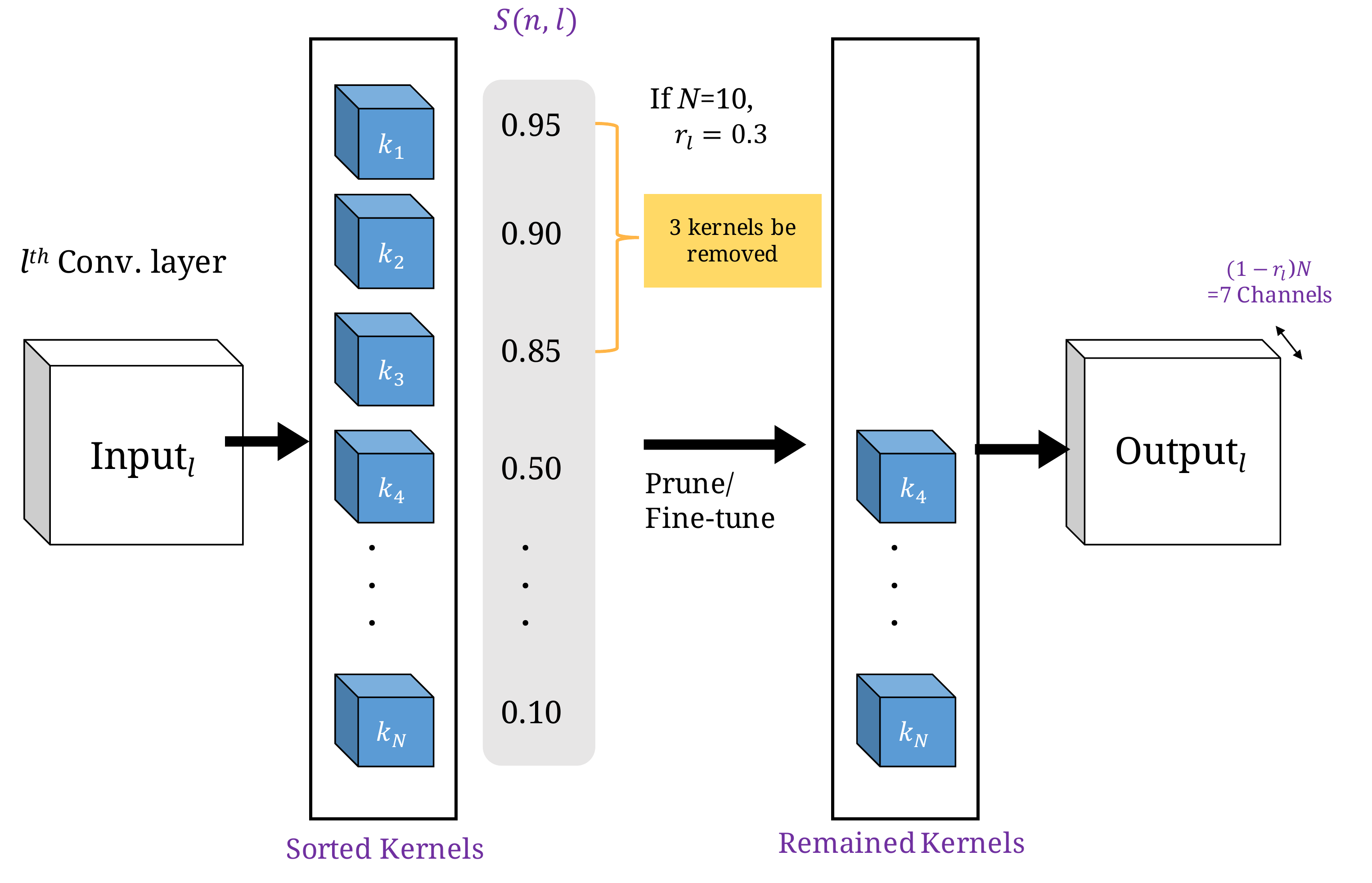}
\caption{Illustration of pruning redundant kernels at $l$-th layer with sorted list of their sparsity. If $N$=$10$ and $r_{l}$=$0.3$ for example, we will remove $r_lN$=$3$ kernels with the highest three sparsity values (0.95, 0.90, 0.85), and the $l$-th output will left $(1-r_l)N$=$7$ channels. }
\label{fig:K}
\end{figure}

\subsection{Layer-Wise Polishment (LWP)}

Given a computation budget, LWP can help find the optimized reducing factor for removing redundant kernels.

After calculating the sparsity of all convolution kernels with (\ref{eq:sparsity}), we sort sparsity values of the same layer $l$ to form the sparsity list $[S_l(1),S_l(2),\dots,S_l(N)]$ in descending order. 
We start to remove the kernels from those with maximum sparsity values, and the number of kernels to be removed of each layer is based on a reducing factor $\mathbf{r} \in [0,1)^{L}$.
The element $r_{l}$ means the proportion of kernels to be removed at the $l$-th layer, and $L$ is the total number of convolution layers. As illustrated in Figure~\ref{fig:K}, if $N$=10, $r_{l}$=0.3 and given the sorted kernels, we will prune $r_lN$=$3$ redundant kernels at the $l$-th layer, and use the remaining kernels to convolve the input feature maps. After pruning the redundant kernels, we will fine-tune the CNN models to reinforce all kernels left; hence, we can still retain a great performance.

Different from Chen's work \cite{chen2016deep}, LWP can determine the reducing factor $\mathbf{r}$ under an expected computation budget. 
This procedure is split into two steps. To begin with, we uniformly increase the elements in $\mathbf{r}$ and use it to prune the kernels, for example from $\mathbf{r}=\mathbf{0.1}$ to $\mathbf{r}=\mathbf{0.6}$, then count the remaining parameters. After fine-tuning every models with different $\mathbf{r}$, we choose one $\mathbf{r}$ = $\mathbf{r}_{\text{fix}}$ with the greatest performance under our computation budget, and then go forward to Step two, which is a key process that can further improve the performance. 

To calculate the parameters remained of all the kernel weights, we can use the following formula. 
\begin{equation}     
    \begin{split}
        weights\ remained(\mathbf{r^{\prime}}) = [1,\ (\mathbf{r}^{\prime}_{{\scriptscriptstyle1:(L-1)}})^{T}]\mathbf{D} \mathbf{r}^{\prime}.
    \end{split} \label{eq:weight_remained}
\end{equation}
The vector $\mathbf{r^{\prime}}=\mathbf{1}-\mathbf{r}$ means the proportion of kernels remained at every layers. $\mathbf{D} \in \mathbb{R}^{L\times L}$ is a diagonal matrix with $D_{ii} = W_{i}H_{i}C_{i}N_{i}$, which is the product of the four dimensions in the 4-D kernel tensor at the $i$-th layer.  $[1,\ (\mathbf{r}^{\prime}_{{\scriptscriptstyle1:(L-1)}})^{T}] \in \mathbb{R}^{1\times L}$ is a vector concatenate $1$ and the slice of $(L-1)$ elements in $\mathbf{r}^{\prime}$. 

Back to Step two, we split the CNN model into three segments, front, middle and end segments, then separately adjust the reducing factor of each part: $\mathbf{r}_{\text{\scriptsize front}}$, $\mathbf{r}_{\text{\scriptsize middle}}$ and $\mathbf{r}_{\text{\scriptsize end}}$.
We explore the idea that assigning the same reducing factors for all layers is not the best way to prune the model; therefore, we can find out the characteristic of a model through experiments and know which segment is more redundant than others. In this paper, we reveal that we can increse the reducing factor in the front segment, and decrease those in both middle and end segments, while maintaining the computation budget almost the same as before, as follows.
\begin{equation}
    \begin{aligned}
        &[1,\ (\mathbf{r}^{\prime}_{\text{fix},1:(L-1)})^{T}]\mathbf{D} \mathbf{r}^{\prime}_{\text{fix}}\approx \\
        &  [1,\ (\mathbf{r}^{\prime}_{\text{\scriptsize front}};\mathbf{r}^{\prime}_{\text{\scriptsize middle}};\mathbf{r}^{\prime}_{\text{\scriptsize end},1:(\mathit{END}-1)})^{T}]\mathbf{D}(\mathbf{r}^{\prime}_{\text{\scriptsize front}};\mathbf{r}^{\prime}_{\text{\scriptsize middle}};\mathbf{r}^{\prime}_{\text{\scriptsize end}}),
        \end{aligned}
\end{equation}
where $\mathbf{r}_{\text{\scriptsize front}}$ = $\mathbf{r}_{\text{\scriptsize fix,front}}+\mathbf{\delta}_{\text {\scriptsize front}}$, $\mathbf{r}_{\text{\scriptsize middle}}$ = $\mathbf{r}_{\text{\scriptsize fix,middle}}-\mathbf{\delta}_{\text{\scriptsize middle}}$ and $\mathbf{r}_{\text{\scriptsize  end}}$ = $\mathbf{r}_{\text{\scriptsize fix,end}}-\mathbf{\delta}_{\text{\scriptsize end}}$. $\mathbf{\delta}$ is the changing vector, which is decided empirically, and $\scriptstyle \mathit{END}$ is the number of layers of the end segment.

 Again, by fine-tuning the model with our adjusted reducing factor, the performance will be further improved compared to that with uniform $\mathbf{r}_{\text{fix}}$. The objective of LWP is trying to keep an expected number of parameters, or we call it computation budget, then regulate the reducing factors of different layers to achieve the best performance.  

\begin{figure}[!t]   
\centering
\includegraphics[width=3.5in]{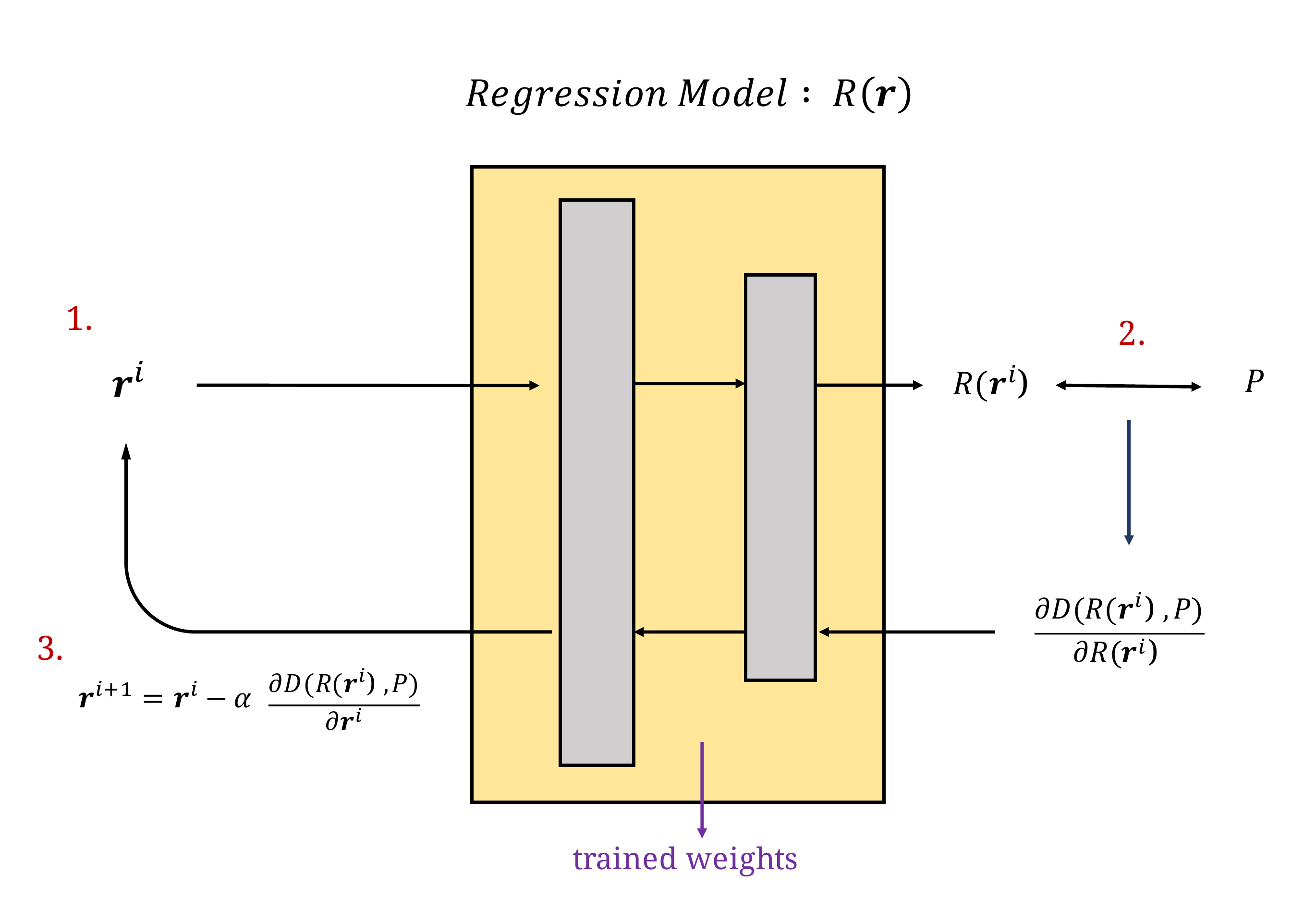}
\caption{Steps of optimizing the reducing factor between every iterations in Gradient Optimization (GO).
Step 1 is feeding the $\mathbf{r}$ to the model. Step 2 is calculating the difference and gradients, then backpropagate. Step 3 is updating the input vector $\mathbf{r}$.}
\label{fig:Re}
\end{figure}

\subsection{Gradient Optimization (GO)}  

From another aspect, given a target performance of the network, GO can help derive the optimized reducing factor $\mathbf{r}$.

This method is based on a learnable regression model, where the input vector is $\mathbf{r}$, and the output is the performance criterion,
whis is PSNR drop in our experiment.
We generate some training data with randomly assigned $\mathbf{r}$ and collect the output performance with fine-tuned CNN model.
With the training data, we construct a simple regression model, $R(\mathbf{r})$, with neural netwok as shown in Figure~\ref{fig:Re}.
After training, we fix the weights inside the model, and use this trained model to start the iteration loops by giving $\mathbf{r}$ an initial point $\mathbf{r}^{0}$.

The iterations operate with a goal to minimize the difference between $R(\mathbf{r})$ and $P$, where $P$ is an expected performance we can claim. After calculating the gradients, they will be backpropagated to the input $\mathbf{r}$, and $\mathbf{r}$ will be updated by gradient descent as the following equation for every iteration:
\begin{equation} 
    \mathbf{r}^{i+1} = \mathbf{r}^{i} - \alpha \frac{\partial \mathit{D}(R(\mathbf{r}^{i}),P)}{\partial \mathbf{r}^{i}},
\end{equation}
where $\alpha$ is the learning rate, $i$ is the iteration index and $\mathit{D}(R(\mathbf{r}),P)$ denotes the difference between $R(\mathbf{r})$ and $P$. These operations are also illustrated in Figure~\ref{fig:Re}. Finally, after some iterations, we can minimize the distance between $P$ and $R(\mathbf{r})$ within an acceptable margin, and the optimal reducing factor $\mathbf{r}_{\text{optimized}}$ for the CNN models is obtained. 

\subsection{Architecture specified tuning}\label{ss:archi} 
Those two methods can both prune the kernels and reduce the computation time. We can either choose one of them to optimize a convolutional neural network according to different application scenarios. 
However, owing to the architecture (instruction set, cache, etc.) of each processor, we further take the architecture characteristics into consideration. For example, in Intel processors, the computation time of CNN inference step is especially fast when the kernels of CNN are certain numbers such as 32. 
The final flows are shown in Procedures LWP and GO as follows.

\begin{algorithm}[h]
  \caption{\textit{Modified Flow for LWP}}
  \begin{algorithmic}[1]
    \State{Choose a trained CNN models and zero the $\mathbf{r}$.}
	\Repeat
	\State{Uniformly increase the elements of $\mathbf{r}$ under\par}
    \Statex{\ \ \ \ \  architecture limitation.}
	\State{Prune and Fine-tune the model.}
	\State{Calculate the parameters remained.}
	\State{Testing the performance compared to the original CNN \phantom a \phantom . \phantom .models.}
	\Until{achieve the computation budget with great \par performance.}
    \State{Save the $\mathbf{r}_{\text{fix}}$ we've choosed.}
    \State{Split $\mathbf{r}_{\text{fix}}$ into three parts, and modify them with considering architecture characteristics.}
    \State{At the end, fine-tune the model with new $\mathbf{r}$.}
  \end{algorithmic}
\end{algorithm}
\begin{algorithm}[!h]
  \caption{\textit{Modified Flow for GO}}
  \begin{algorithmic}[1]
    \State{Choose a trained CNN models.}
	\Repeat
    \State{Randomly generate $\mathbf{r}$.}
	\State{Prune and Fine-tune the model, testing the\par performance $P$. }
    \State{Collect training pair ($\mathbf{r}$,$P$).}
	\Until{Enough training data.}
    \State{Train a regression model $R(\mathbf{r})$ whose input vector is $\mathbf{r}$, output number is $P$.}
    \State{Fix weights inside $R(\mathbf{r})$.}
    \State{Claim an expected $P$, and initiate the start point $\mathbf{r}^{0}$.}
    \Repeat
    \State{Minimizing the difference $D(R(\mathbf{r}),P)$ by updating $\mathbf{r}^{0}$.}
    \Until{$D$ less than a margin at the $i^{th}$ iteration.}
    \State{Slightly modified $\mathbf{r}_{\text{optimized}}$ with considering architecture characteristics.}
    \State{At the end, fine-tune the model with new $\mathbf{r}$.}
  \end{algorithmic}
\end{algorithm}

\section{Experiment Result}
\begin{figure}[!t]  
\centering
\includegraphics[width=0.48\textwidth]{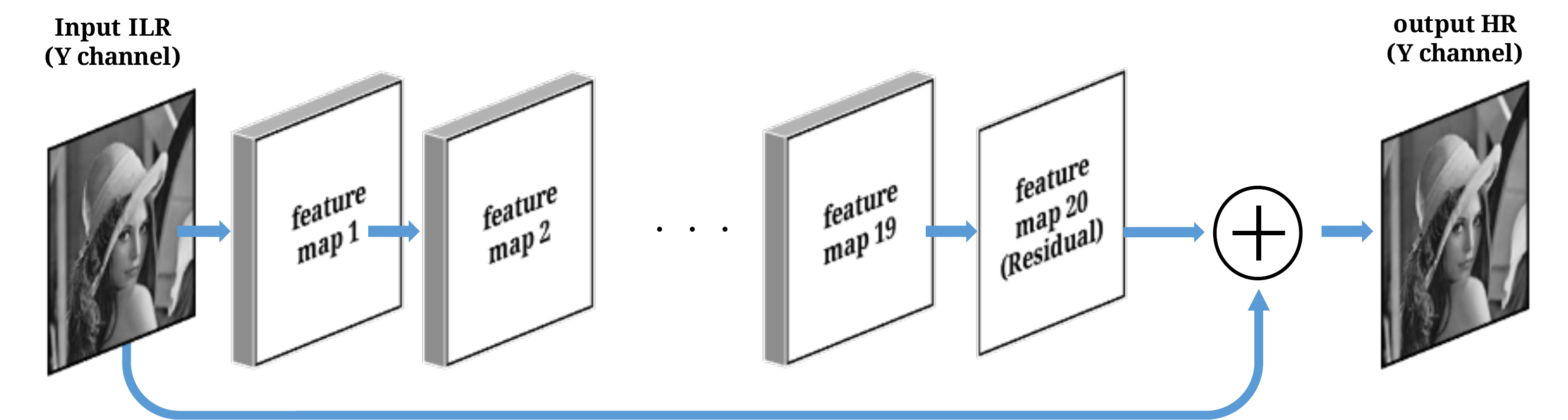}
\caption{Network structure of VDSR. It cascades a pair of layers (convolutional and nonlinear) repeatedly for 20 times. Input is the Y channel of an Interpolated Low-Resolution (ILR) image, and then transforms into the Y channel of the High-Resolution (HR) one. The network predicts a residual image, and the addition of the input and the residual gives the desired output. }
\label{fig:V}
\end{figure}

We take the residual CNN model in Very Deep Super Resolution \cite{kim2016accurate} (VDSR) for our experiment. This model is constructed only with convolutional layers; therefore, the model size and the computation time will not be influenced by the fully-connected layers. Figure~\ref{fig:V} is the VDSR model \cite{kim2016accurate}. The input image is an interpolated low-resolution picture with one channel (Y channel), and the output is the derived high-resolution one. The structure is a 20-layer residual CNN model. Each of the first 19 layers has 64 kernels, and one kernel at the last layer to generate the residual part, which is added by the low resolution image to generate the high-resolution one.



\subsection{Experiment of Layer-Wise Polishment (LWP)}
According to the procedure of LWP, we choose seven different $\mathbf{r}$ for experiments as shown in Table~\ref{table:r}. After pruning, we fine-tune the VDSR model with only three epochs, which is much lower than the training step of the original VDSR network, which is about 80 epochs. Our testing set are Set5$\times2$ and Set14$\times2$ of SR, and the performance criterion we choose is the PSNR drop compared to those with the original VDSR network. Parameters remained are calculated by (\ref{eq:weight_remained}) and the experimental results are shown in Table~\ref{table:r}. We then choose $\mathbf{r}_{\text{fix}}$ = $\mathbf{0.25}$ to be modified later because almost $50\%$ parameters are removed while the performance is still acceptable, where the PSNR drop is only $0.29$ dB.

\begin{table}[t]  
\centering
\caption{LWP: Fine-tune Result (3 epochs) of Different Reducing Factor. \hspace{\textwidth}(The reduce factor number
is chosen by architecture characteristic.)}
\label{table:r}
\resizebox{0.5\textwidth}{!}{
\begin{tabular}{|c|c|c|c|c|}
\hline
& \textbf{\begin{tabular}[c]{@{}c@{}}Reduce\\ Factor $\mathbf{r}$\end{tabular}} & \textbf{\begin{tabular}[c]{@{}c@{}}Kernel\\ per layer\end{tabular}} & \textbf{\begin{tabular}[c]{@{}c@{}}Weights\\ remained (\%)\end{tabular}} & \textbf{\begin{tabular}[c]{@{}c@{}}PSNR drop\\ Set5$\times2$ /Set14$\times2$ \end{tabular}} \\ \hline\hline
    \textbf{\begin{tabular}[c]{@{}c@{}}Original\\ VDSR\end{tabular}} & $\mathbf{0}$ & 64 & 100\% & \begin{tabular}[c]{@{}c@{}}0/0\\ (PSNR: 37.50 / 33.08)\end{tabular} \\ \hline
        \multirow{7}{*}{\textbf{\begin{tabular}[c]{@{}c@{}}Experiment\\ Result\end{tabular}}} & $\mathbf{0.12}$ & 56 & 76.6\% & 0.16 / 0.20 \\ \cline{2-5} 
 & $\mathbf{0.18}$ & 52 & 66.1\% & 0.37 / 0.29 \\ \cline{2-5} 
 &\cellcolor[HTML]{EFEFEF}$\mathbf{0.25}$ & \cellcolor[HTML]{EFEFEF}48 & \cellcolor[HTML]{EFEFEF}56.3\% & \cellcolor[HTML]{EFEFEF}0.29 / 0.27 \\ \cline{2-5} 
 & $\mathbf{0.32}$ & 44 & 47.4\% &  0.42 / 0.40 \\ \cline{2-5} 
 & $\mathbf{0.38}$ & 40 & 39.2\% & 0.49 / 0.47  \\ \cline{2-5} 
 & $\mathbf{0.44}$ & 36 & 31.7\% &  0.50 / 0.48\\ \cline{2-5} 
 & $\mathbf{0.50}$ & 32 & 25.1\% &  0.65 / 0.56 \\ \hline
\end{tabular}
}
\end{table}

We then split the 20-layers CNN model into three segments with 6, 7 and 7 layers respectively, and make an experiment for increasing and decreasing the reducing factor for each segment and try to find out the characteristic of the deep CNN model. As Table~\ref{table:parts} suggests, under almost the same parameters remained, we can improve the performance: for Set5$\times2$, the PSNR drop is decreased from $0.29$ to $0.24$ when pruning more kernels at the front segment ($\mathbf{r}_{\text{front}}= \mathbf{0.44}$) and retaining more at the middle and end segments. 

\begin{table}[t]
\centering
\caption{LWP: Experiment of Different Reducing Factor at Different Fragments. (F,M,E) Means Front, Middle and End Segments.}
\label{table:parts}
\resizebox{0.5\textwidth}{!}{
\begin{tabular}{|c|c|c|c|}
\hline
\textbf{\begin{tabular}[c]{@{}c@{}}Reduce \\ Factor $\mathbf{r}$ \\ (F,M,E)\end{tabular}} & \textbf{\begin{tabular}[c]{@{}c@{}}Kernels \\per part\\ (F,M,E)\end{tabular}} & \textbf{\begin{tabular}[c]{@{}c@{}}Weights\\ remained (\%)\end{tabular}} & \textbf{\begin{tabular}[c]{@{}c@{}}PSNR drop\\ Set5$\times2$ / Set14$\times2$\end{tabular}} \\ \hline\hline
$(0.25,0.25,0.25)$& 48, 48, 48 & 56.3\% & 0.29 / 0.27 \\ \hline
\cellcolor[HTML]{EFEFEF}$(\mathbf{0.44},0.18,0.18)$& \cellcolor[HTML]{EFEFEF}\textbf{36}, 52, 52 & \cellcolor[HTML]{EFEFEF}55.4\% & \cellcolor[HTML]{EFEFEF}0.27 / 0.26 \\ \hline
\cellcolor[HTML]{EFEFEF}$(\mathbf{0.44},0.12,0.25)$& \cellcolor[HTML]{EFEFEF}\textbf{36}, 56, 48 & \cellcolor[HTML]{EFEFEF}56.4\% & \cellcolor[HTML]{EFEFEF}0.24 / 0.26 \\ \hline
$(0.12,0.18,0.44)$& 56, 52, 36 & 58.6\% & 0.34 / 0.33 \\ \hline
$(0.18,0.18,0.38)$& 52, 52, 40 & 57.7\% & 0.38 / 0.33 \\ \hline
$(0.25,0.44,0.06)$& 48, 36, 60 & 55.9\% & 0.37 / 0.38 \\ \hline
$(0.18,0.44,0.12)$& 52, 36, 56 & 55.5\% & 0.41 / 0.40 \\ \hline
\end{tabular}
}
\end{table}

\subsection{Experiment of Gradient Optimization (GO)}
In the procedure of GO, we first need to collect enough training data pairs ($\mathbf{r}$, $P$), where $P$ we use here is the PSNR drop when testing on Set14$\times2$ with randomly assigned $\mathbf{r}$. Then we construct a linear regression model $R(\mathbf{r})$ whose input vector is $\mathbf{r}$ and output is $P$. The training criterion is mean-square-error (MSE) loss, and the optimizer is Adam Optimizer \cite{DBLP:journals/corr/KingmaB14}. After some iterations, we can get the well-trained $R(\mathbf{r})$.

The trained regression model is then employed to optimize the input vector $\mathbf{r}$.
With this $R(\mathbf{r})$, GO can automatically decide the appropriate reducing factor $\mathbf{r}$ to prune our CNN model under the expected performance. We use $P=0.29$ as an example, as shown in Table~\ref{table:methodb}.
The modified $\mathbf{r}$ will be used to prune our model, and Set5$\times2$ and Set14$\times2$ are used as the testing sets. We first give $\mathbf{r}$ an initial point $\mathbf{r}^{0}$ and use the absolute difference $\abs{P-R(\mathbf{r})}$ as the loss $D(R(\mathbf{r}),P)$. Between every iteration, we calculate the gradient of loss and backpropagate it to optimize input $\mathbf{r}$ with Gradient Descent (learning rate $\alpha= 0.01$). After minimizing $D$ within a margin, we stop the iteration and get the optimized $\mathbf{r}$, and then start to adjust it with considering the architecture characteristics.
The result of the modified reducing factor and the performance after fine-tuning the CNN model are also shown on Table~\ref{table:methodb}.
We can clearly see that the final PSNR drop results (0.24dB/0.25dB) are successfully close to the goal we set.

\begin{table}[!t]
\centering
\caption{GO: Experiment Result of Goal = 0.29 PSNR drop.}
\label{table:methodb}
\resizebox{0.49\textwidth}{!}{
\begin{tabular}{|c|c|}
\hline
\multicolumn{2}{|c|}{\rule{0pt}{9pt}\begin{tabular}[c]{cc@{}c@{}}\textbf{\small Goal: 0.29 PSNR drop}\end{tabular}}\\ \hline\hline
\multirow{2}{*}{\begin{tabular}[c]{@{}c@{}}Optimized by\\ Gradient Descent\end{tabular}} & \rule{0pt}{9pt}\textbf{Reducing Factor} $\mathbf{r}$ \\ \cline{2-2} 
    &\rule{0pt}{9pt} [0.30, 0.07, 0.12, 0.07, 0.11,$...$, 0.27, 0.08, 0.08, 0.18, 0]$^{T}$\\ \hline\hline
\multirow{4}{*}{\begin{tabular}[c]{@{}c@{}}Adjusted by\\ Architecture\\ Characteristics\end{tabular}} &\rule{0pt}{9pt} \textbf{Reducing Factor} $\mathbf{r}$ \\ \cline{2-2} 
    & \rule{0pt}{9pt} [0.32, 0.06, 0.12, 0.06, 0.12,$...$, 0.25, 0.12, 0.12, 0.18, 0]$^{T}$ \\ \cline{2-2} 
 & \rule{0pt}{8pt}\textbf{\# Kernel Remained}\ (20 layers) \\ \cline{2-2} 
 & \rule{0pt}{9pt}(44, 60, 56, 60, 56,$...$, 48, 56, 56, 52, 1) \\ \hline
\begin{tabular}[c]{@{}c@{}}Performance\\ (PSNR drop)\end{tabular}& \begin{tabular}[c]{@{}c@{}}Set5($\times 2$) / Set14($\times 2$)\\\textbf{0.29 / 0.26} \end{tabular} \\ \hline
\end{tabular}
}
\end{table}
\section{Conclusion}  
In this paper, we propose computation-performance optimization (CPO) methods with removing the redundant kernels of a deep convolutional neural network (CNN) to make it more feasible on embedded systems. Two CPO methods, LWP and GO, are proposed. The first achieves about $50\%$ size reduction but only causes about 0.2dB in performance drop. The other can automatically derive a removing policy under our performance goal. Compared to the previous works about kernel pruning, our work is more flexible than that with a fixed threshold and can be applied to more complex network models. In the future, we expect to profile more networks and apply it to even more architectures, such as ASIC or FPGA platforms.


\section*{Acknowledgment}
This research was supported in part by the Ministry of Science and Technology of Taiwan (MOST 107-2633-E-002-001), National Taiwan University, Intel Corporation, and Delta Electronics.




\bibliographystyle{IEEEtran}
\bibliography{IEEEabrv,jackie}

\end{document}